\documentclass{article}

\usepackage{times}
\usepackage{graphicx} % more modern
\usepackage{subfigure} 
\usepackage{enumerate}
\usepackage{natbib}

\usepackage{algorithm}
\usepackage{algorithmic}

\usepackage{hyperref}

\usepackage[accepted]{icml2016}

\icmltitlerunning{Simple and Efficient Tensor Regression}

\usepackage{tablefootnote}

\newcommand{\loss} {\mathcal{L}}

\newcommand{\E}{{\mathcal E}}

\newcommand{\W}{{\mathcal W}}
\newcommand{\R}{{\mathbb R}}   %% mathbb not working?
\newcommand{\X}{{\mathcal X}}
\newcommand{\Y}{{\mathcal Y}}

\newcommand{\skt}{{\mathcal{S}}}

\usepackage{amsmath,amssymb,amsthm,bm}
\DeclareMathOperator*{\argmax}{argmax}
\DeclareMathOperator*{\argmin}{argmin}
\newcommand{\eat}[1]{}

\newtheorem{theorem}{Theorem}[section]

\newtheorem{definition}[theorem]{Definition}
\newtheorem{lemma}[theorem]{Lemma}

\newcommand{\M}[1]{{\mathbf{#1}}} % matrix
\newcommand{\T}[1]{{\mathcal{#1}}} % tensor
\newcommand{\V}[1]{{\mathbf{#1}}} % vector

\newcommand{\uu}{\M{U}}

\usepackage{color}

\begin{document} 

\twocolumn[
\icmltitle{ Learning from Multiway Data: Simple and Efficient Tensor Regression   }

\icmlauthor{Rose Yu}{qiyu@usc.edu}
\icmlauthor{Yan Liu}{yanliu.cs@usc.edu}
\icmladdress{Department of Computer Science, University of Southern California}

\icmlkeywords{tensor, regression, multi-way data, spatio-temporal analysis}

\vskip 0.3in
]

\begin{abstract} 
%Large-scale multivariate spatio-temporal data are ubiquitous. One way of accelerating the process of spatio-temporal data analysis is via downsampling the observations. Many tasks in multivariate spatio-temporal data analysis can be formulated as  low-rank tensor learning problem, such as time series forecasting.  In this paper, we propose a low-rank tensor sketching algorithm for  spatio-temporal sampling problem. Specifically, we generalize the recent development in sparse subspace embedding to low-rank tensor domain. We show that our algorithm achieves accurate predictions with significant speed-up in social media and climate applications.

%Tensor regression has shown to be a powerful tool for analyzing spatio-temporal data  by exploiting the  complex structure of tensor covariants. However, 

Tensor regression has shown to be advantageous in learning  tasks with multi-directional relatedness. Given massive multiway data, traditional methods  are often too slow to operate on or suffer from memory bottleneck. In this paper, we introduce subsampled  tensor projected gradient   to  solve the problem.  Our algorithm is impressively simple and efficient. It is built upon  projected gradient method with fast tensor power iterations,  leveraging randomized  sketching  for further acceleration.  Theoretical analysis shows that our algorithm converges to the correct solution in fixed number of iterations. The memory requirement grows linearly with the size of the problem.   We  demonstrate  superior empirical performance on  both multi-linear multi-task learning and spatio-temporal  applications. 

\end{abstract} 

\section{Introduction}
\label{introduction}
%introduction
% what is the problem
Massive multiway  data emerge from many fields: space-time measurements on  several variables in  climate dynamics \cite{hasselmann1997multi}, multichannel EEG signals in neurology  \cite{acar2007multiway} and  natural images sequences in computer vision \cite{vasilescu2002multilinear}.  Tensor provides a natural representation for multiway data. In particular, tensor decomposition has been a popular technique for exploratory data analysis \cite{kolda2009tensor} and has been extensively studied.  In contrast, tensor regression, which aims to learn a model with multi-linear parameters, is especially suitable for applications with multi-directional relatedness, but has not been fully examined. For example, in a task that predicts multiple climate variables at different locations and time, the data can be indexed by \textit{variable} $\times$ \textit{location} $\times$ \textit{time}. Tensor regression provides us with a concise way of modeling complex structures in multiway data.

Many tensor regression methods have been proposed \cite{zhao2011multilinear, zhou2013tensor, romera2013multilinear, wimalawarne2014multitask, signoretto2014learning}, leading to a broad range of successful applications ranging from neural science to climate research to social network analysis. These methods share the  assumption that the model parameters form a high order tensor and there exists  a low-dimensional factorization for the model tensor. They can be summarized into two types of approaches: (1) alternating least square (ALS) sequentially finds the factor that minimizes the loss while keeping others fixed; (2) spectral regularization approximates the  original non-convex problem with a convex surrogate loss, such as the nuclear norm of the unfolded tensor.

A clear drawback of all the algorithms mentioned above is high computational cost.  ALS displays unstable convergence properties  and outputs sub-optimal solutions \cite{cichocki2009nonnegative}.  Trace-norm minimization suffers from slow convergence \cite{gandy2011tensor}. Moreover, those methods face the memory bottleneck when dealing with large-scale datasets. ALS, for example, requires the matricization of entire data tensor at every mode. In addition, most existing algorithms are largely constrained by the specific tensor regression model.  Adapting one algorithm to a new regression model involves derivations for all the updating steps, which can be tedious and sometimes non-trivial.

% how other people solve it
In this paper, we  introduce subsampled  Tensor Projected Gradient (TPG), a simple and fast recipe to address the challenge.  It is an efficient solver for a variety of tensor regression problems. The memory requirement grows linearly with the size of the problem.   Our  algorithm is  based upon  projected gradient descent \cite{calamai1987projected} and can also be seen as a tensor generalization of iterative hard thresholding  algorithm \cite{blumensath2009iterative}. At each projection step, our algorithm iteratively returns a set of  leading singular vectors of the model,  avoiding  full singular value decomposition (SVD).  To handle large sample size, we  employ randomized sketching for subsampling and noise reduction to further accelerate the process. 

We provide theoretical analysis of our algorithm, which  is guaranteed to find the correct solution under the Restricted Isometry Property  (RIP) assumption.  In fact, the algorithm only needs a fixed number of iterations, depending solely on the logarithm of  signal to noise ratio. It is also robust to noise, with  estimation error depending linearly on the size of the observation error.  The proposed method is simple and easy to implement.  At the same time, it enjoys fast  convergence  rate and superior robustness.  We  demonstrate the empirical  performance on two example applications: multi-linear multi-task learning and multivariate spatio-temporal forecasting. Experiment results show that the proposed  algorithm significantly outperforms existing approaches in  both prediction accuracy and speed.

\begin{table*}[h]
	\caption{Summarization of contemporary tensor regression models,  algorithms and applications} %title of the table
	\begin{center}
		%\begin{sc}
		\begin{tabular}{c|c|c}
			\hline 
			MODEL  & ALGO &  APP \\
			\hline	
			$\T{Y}= \text{cov} \langle \X , \W \rangle + \E $ \tablefootnote{$\text{cov}(\X, \W):\T{Y}_{\cdots,i_{n-1}i_{n+1}\cdots ,j_{n-1}j_{n+1},\cdots }= \sum_{i_n}\X_{i_n}\W_{i_n}$}  \cite{zhao2011multilinear}  & High-order Partial Least Square & Neural Imaging (EEG)\\
						\hline
			$\Y = \text{vec} (\X)^T \text{vec}(\W)  + \E $  \cite{zhou2013tensor} & Alternating Least Square & Neural Imaging (MRI)\\
			\hline
			$\M{Y}^t =  \M{X}^t \V{w }^t + \V{\epsilon}^t$  \cite{romera2013multilinear} & ADMM & Multi-task learning  \\
			\hline
			$\M{Y}_m =  \T{X}_m \M{W}_m + \M{E}_m$ \cite{yu2014fast}   & Orthogonal Matching Pursuit & Spatio-temporal Forecasting\\
			\hline
			%	$\Y = \X \times_1 \M{W}_{1} \cdots \times_n \M{W}_n +\E$  \cite{hoff2015multilinear} & Bayesian Estimation & Social Networks \\
			%	\hline 
		\end{tabular}
		%\end{sc}
	\end{center}	
	\label{tb:tensor_model}	
\end{table*}

\section{Preliminary}
\label{prelim}
Across the paper, we use calligraphy font for tensors, such as $\T{X},\T{Y}$, bold uppercase letters for matrices, such as $\M{X},\M{Y}$, and bold lowercase letters for vectors, such as $\V{x},\V{y}$.

\textbf{Tensor Unfolding}
Each dimension of a tensor is a mode. An n-mode unfolding of a tensor $\W$  along mode $n$ transforms a tensor into a matrix $\W_{(n)}$ by treating $n$ as the first mode of the matrix and cyclically concatenating other modes. The indexing follows the convention in \cite{kolda2009tensor}. It is also known as tensor matricization. 
%Multiple mode unfolding is not considered in this paper \ryedit{We only consider single mode unfolding or Remove this sentence}.

\textbf{N-Mode Product}
The n-mode product between tensor $\W$ and matrix $\uu$ on mode $n$ is represented as $\W\times_n \M{U}$ and is defined as $(\W\times_n\M{U})_{(n)} = \M{U} \W_{(n)}$ .

% \textbf{reduced QR decomposition}
% QR decomposition factorize the matrix $W\in \R^{m\times n}$ into a product of a unitary matrix $Q \in \R^{m\times m}$ and an upper triangular matrix $R \in \R^{m\times n}$, $m \geq n$. the reduced QR composition of order $r$ consists of the first $r$ columns of $Q$ and top $r$ rows of R.

\textbf{Tucker Decomposition}
Tucker decomposition factorizes a tensor $\W$ into 
$\W = \skt \times_1 \M{U}_1 \cdots \times_n \M{U}_n$, where $\{\M{U}_n\}$ are all unitary matrices and the core tensor satisfies that $\skt_{(n)}$ is row-wise orthogonal for all $n=1,2,\dots,N$.

\section{Related Work}
\label{relate}
%related work
% why it is challenging
Several algorithms have  been proposed for tensor regression. For example,  \cite{zhou2013tensor}  proposes  to use  Alternating least square (ALS) algorithm. \cite{romera2013multilinear}  employs ALS as well as an Alternating Direction Method of Multiplier (ADMM) technique  to  solve the nuclear norm regularized optimization problem.   \cite{signoretto2014learning} proposes a more general version of ADMM  based on Douglas-Rachford splitting method. Both ADMM-based algorithms try to solve a  convex relaxation of the original optimization problem, using singular value soft-thresholding. To address the scalability issue of these methods,  \cite{yu2014fast} proposes a greedy algorithm following  the Orthogonal Matching Pursuit (OMP) scheme. Though significantly faster, it requires the matricization of the data tensor, and thus would  face memory bottleneck  when dealing with large sample size.

%\cite{hoff2015multilinear} proposed a Bayesian variant of the tensor regression model and solves with posterior estimation. 

% summarize contribution

Our work is closely related to  iterative hard thresholding  in compressive sensing \cite{blumensath2009iterative}, sparsified gradient descent in sparse recovery  \cite{garg2009gradient} or singular value projection method in  low-rank matrix completion \cite{jain2010guaranteed}. We generalize the idea of iterative hard thresholding to tensors and utilize  several tensor specific properties to achieve acceleration. We also leverage randomized sampling technique, which concerns  how to sample data to accelerate the common learning algorithms. Specifically, we employ count sketch  \cite{clarkson2013low} as a pre-processing step to alleviate the memory bottleneck for large dataset.

\section{Simple and Efficient Tensor Regression}
\label{method}
% Introduction to the problem 
We start by describing the problem of tensor regression and our proposed algorithm in details. We use three-mode tensor for ease of explanation. Our method and analysis directly applies to higher order cases.

\subsection{Tensor Regression}

 Given a predictor tensor $\X $ and a response tensor $\Y $, tensor regression targets at the following problem:
\begin{eqnarray}
	\label{eqn:tensor_regression}
	\W^\star = \argmin\limits_{\W} \loss(\W ; \X, \Y)   \nonumber \\
 \text{s.t.} \quad  \text{rank} (\W) \leq R 
\end{eqnarray}
 The problem aims to estimate a  model tensor  $\W\in \mathbb{R}^{ D_1 \times D_2 \times D_3}$  that minimizes the empirical loss $\loss$, subject to the constraint that the Tucker rank of $\W$ is at most $R$. Equivalent, the model tensor $\W$ has a low-dimensional factorization $\W = \T{S} \times_1 \M{U}_1 \times_2 \M{U}_2\times_3\M{U}_3$ with core $\T{S}\in \R^{R_1 \times R_2 \times R_3}$ and orthonormal projection matrices $\{\M{U}_n \in \R^{D_n \times R_n} \}$. The dimensionality of $\T{S}$ is at most $R$. The reason we favor Tucker rank over others is due to the fact that it is a high order generalization of matrix SVD, thus is computational tractable, and carries nice properties that we later would utilize.

Many  existing tensor regression models are special cases of the problem in (\ref{eqn:tensor_regression}). For example, in multi-linear multi-task learning \cite{romera2013multilinear}, given the predictor and response for each task $(\M{X}^t, \M{Y}^t)$,  the empirical loss is defined as the summarization of the loss for all the tasks, i.e., $\loss(\W; \X, \Y) = \sum_{t=1}^T \| \M{Y}^t -  \M{X}^t \V{w }^t \|^2_F$, with $\V{w}^t$ as the $t$th column of $\W_{(1)}$.  For the univariate GLM model in \cite{zhou2013tensor}, the model is defined as $\Y = \text{vec} (\X)^T \text{vec}(\W)  + \E $.  Table \ref{tb:tensor_model} summarizes existing tensor regression models,  their algorithms as well as  main application domains. In this work, we use the  simple linear regression model $\Y= \langle \X,\W  \rangle  +\E$ to illustrate our idea, where $ \X  \in \mathbb{R}^{T \times D_1  \times D_3 }$, $\Y \in \mathbb{R}^{T \times D_2 \times D_3 }$ with sample size $T$, and $\E $ as i.i.d Gaussian noise.    The tensor inner product  $\langle \X, \W \rangle $ is defined as the  matrix multiplication on each slice independently, i.e., $\langle \X, \W \rangle =\sum_{m=1}^M \X_{:,:,m}, \W_{:,:,m} $. Our methodology can be easily extended to handle more complex regression models.

\subsection{ Tensor Projected Gradient}
To solve  problem (\ref{eqn:tensor_regression}), we propose a simple and efficient tensor regression algorithm: subsampled Tensor Projected Gradient (TPG). TPG is based upon the prototypical method of projected gradient descent  and can also be seen as a tensor generalization of iterative hard thresholding  algorithm \cite{blumensath2009iterative}.  For the  projection step, we  resort to tensor power iterations to iteratively search for the leading singular vectors of the model tensor.  We further leverage  randomized sketching \cite{clarkson2013low} to  address the memory bottleneck  and speed up the algorithm.

%Alternatively, the model can also be written in matrix form as  $\text{vec}(\T{Y}) = \text{diag}(\T{X}) \otimes \M{I}_P \text{vec}(\T{W})+  \text{vec}(\epsilon) $. This shows that the tensor inner product is a linear operator. 
 
As shown in Algorithm \ref{alg:TPG}, subsampled Tensor Projected Gradient (TPG)  combines a gradient  step with a proximal point  projection step  \cite{rockafellar1976monotone}. The gradient step treats (\ref{eqn:tensor_regression}) as an unconstrained optimization of $\W$. As long as the loss function  is differentiable  in a neighborhood of  current solution,  standard gradient descent methods can be applied.  For our case, computing the gradient under linear model is trivial: $\triangledown \T{L}(\W; \X, \Y) = \langle \X^T, \Y-\langle \X, \W\rangle \rangle$.  After the gradient step, the subsequent  proximal point  step aims to find a projection $\T{P}_R (\W):  \R^{D_1\times D_2 \times D_3} \rightarrow  \R^{D_1\times D_2 \times D_3}$ satisfying: 
 \begin{eqnarray}
\T{P}_R(\W^{k}) =	\argmin\limits_{\W} (\|\W^k -  \W\| _F^2) \nonumber \\
 	 \text{s.t.} \quad  \W \in \T{C}(R) = \{\W: \text{rank} (\W) \leq R\}
 \end{eqnarray} 
The difficulty of solving the above problem mainly comes from the non-convexity of the set of low-rank tensors.  A common approach  is to use nuclear norm as a convex surrogate  to approximate the rank constraint \cite{gandy2011tensor}. The resulting problem can either be solved by Semi-Definite Programming (SDP) or soft-thresholding of the singular values. However, both algorithms are computational expensive. Soft-thresholding, for example, requires a full SVD for each unfolding of the tensor.

\begin{algorithm}[h]
	\caption{ Subsampled Tensor Projected Gradient }
	\label{alg:TPG}
	\begin{algorithmic}[1]
		\STATE {\bfseries Input:}
		predictor  $ \X$, response  $\Y $, rank $R$
		\STATE {\bfseries Output:} model tensor $ \W \in \mathbb{R}^{ D_1\times D_2 \times D_3}$ 
		\STATE Compute count sketch   $ \M{S}$
		\STATE Sketch $\tilde{\Y} \gets \Y\times_1 \M{S} $, $\tilde{\X} \gets \X\times_1 \M{S}$
		\STATE Initialize 	$\W^0$ as zero tensor
		\REPEAT
		\STATE  $\tilde{\W}^{k+1} = \W^k - \eta \triangledown \T{L}(\W^k; \tilde{\X},\tilde{ \Y})$
		\STATE $\W^{k+1} =   \text{ITP} (\tilde{\W}^{k+1} )$
		\UNTIL{Converge}
	\end{algorithmic}
\end{algorithm}

\begin{algorithm}[h]
	\caption{ Iterative Tensor Projection (ITP) }
	\label{alg:ITP}
	\begin{algorithmic}[1]
		\STATE {\bfseries Input:}
		 model $ \tilde{\W }$,  predictor $\X $, response  $\Y $, rank $R$
		\STATE {\bfseries Output:} projection $ \W \in \mathbb{R}^{D_1\times D_2 \times D_3}$ 
		\STATE Initialize $\{\M{U}_n^0\}$ with  $R$ left singular vectors of $\W_{(n)}$
		\WHILE {$i \leq R$ }
		\REPEAT 
		\STATE $\V{u}_{1}^{k+1}  \gets  \tilde{\W }  \times_2 {\V{u}_{2}^k}^T \times_3 {\V{u}_{3}^k}^T $
		\STATE $\V{u}_{2}^{k+1}  \gets \tilde{\W }\times_1 {\V{u}_{1}^k}^T  \times_3 {\V{u}_{3}^k}^T$
		\STATE $\V{u}_{3}^{k+1}  \gets   \tilde{\W } \times_1 {\V{u}_{1}^k}^T  \times_2 {\V{u}_{2}^k}^T $
		\UNTIL{Converge to $\{\V{u}_1, \V{u}_2, \V{u}_3 \}$}
		\STATE{Update $\{\M{U}_{n}\}$ with $\{\V{u}_n\}$}
		\STATE  $ \W \gets   \tilde{\W } \times_1 \M{U}_1 {\M{U}_1}^T \times_2 \M{U}_2{\M{U}_2}^T \times_3 \M{U}_3 {\M{U}_3}^T  $
		\IF {$\loss(\W; \X, \Y) \leq \epsilon$ }
		\STATE{RETURN}
		\ENDIF
		\ENDWHILE
	\end{algorithmic}
\end{algorithm}

Iterative hard thresholding, on the other hand, avoids full SVD.  It takes advantage of the general Eckart-Young-Mirsky theorem \cite{eckart1936approximation}  for matrices, which allows the Euclidean projection to  be efficiently computed with thin SVD.  Iterative hard thresholding algorithm has been shown to be memory efficient and robust to noise.  Unfortunately, it is well-known that Eckart-Young-Mirsky theorem no long applies to higher order tensors \cite{kolda2009tensor}. Therefore,  computing high-order singular value decomposition  (HOSVD) \cite{de2000multilinear}  and discarding small singular values do not guarantee optimality of the projection. 

To address the challenge, we note that for tensor   Tucker model, we have :  $\W = \T{S} \times_1 \M{U}_1 \times_2 \M{U}_2\times_3 \M{U}_3$. And the projection matrices $\{\M{U}_n\}$ happen to be the left singular vectors of the unfolded tensor, i.e.,  $\M{U}_n \M{\Sigma}_n \M{V}_n^T = \W_{(n)}$. This property allows us to compute each projection matrix efficiently with thin SVD. By iterating over all factors,  we can obtain a local optimal solution that is guaranteed to have rank at most $R$. We want to emphasize that there is no known algorithm that can guarantee the convergence to the global optimal solution. However,  in the Tucker model, different local optimas are highly concentrated, thus the choice of local optima does not really matter \cite{ishteva2011tucker}.

When the model parameter tensor $\W$ is very large, performing thin SVD itself can be expensive. In our problem, the dimensionality of the model is usually much larger than its  rank.  With this observation, we utilize another property of Tucker model $\M{U}_n = \W \times_1\cdots  \times_{n-1} \M{U}_{n-1}^T \times_{n+1} \M{U}_{n+1}^T \cdots \times_N\M{U}_N $. This property implies that instead of performing thin SVD on the original tensor, we can trade cheap tensor matrix multiplication to avoid  expensive large matrix SVD. This leads to the  Iterative Tensor Projection (ITP) procedure as described in Algorithm \ref{alg:ITP}. Denote $\{\V{u}_n \}$ as row vectors of $\{\M{U}_n \}$, ITP uses power iteration to find one leading singular vector at a time. The algorithm stops either when hitting the rank upper bound $R$  or when the loss function value decreases below a threshold $\epsilon$.

ITP is significantly faster than traditional tensor regression algorithms especially when the model is low-rank.  It guarantees that the proximal point projection step  can be solved efficiently. If we initialize our solution with the top $R$ left singular vectors of tensor unfoldings, the projection iteration can start from a close neighborhood  of the stationary point, thus leading to faster convergence.  In tensor regression, our main focus is to minimize the empirical loss. Sequentially finding the rank-$1$ subspace allows us to evaluate the performance as the algorithm proceeds. The decrease of empirical loss would call for early stop of the thin SVD procedure. 

Another acceleration trick we employ is randomized sketching. This trick is particularly useful when we are encountered with ultra high sample size or extremely sparse data.  Online algorithms,  such as stochastic gradient descent or  stochastic ADMM are common techniques to deal with large samples and break the memory bottleneck. However, from a subsampling point of view, online algorithms make i.i.d assumptions of the data and uniformly select samples.  It usually  fails to leverage the data sparsity. 

In our framework, the convergence of the TPG algorithm, as will be discussed in a later section,  depends only on the logarithm of signal to noise ratio. Randomized sketching instantiates the mechanism to reduce noise by duplicating data instances and combining the outputs. This mechanism provides TPG with considerable amount of  boost. Its performance therefore increases linearly if the noise is decreased. We resort to  \textit{count sketch} \cite{clarkson2013low} as a subsampling step before feeding data into  TPG.   A count  sketch matrix  of size $M \times N$, denoted as $\M{S}$, is generated as follows: start with a zero matrix $\mathbb{R}^{M \times N} $ , for each column $j$, uniformly pick a row $i \in \{ 1,2,\cdots M \}$ and  assign $\{-1,1\}$ with equal probability to $\M{S}_{i,j}$.  In practice, we  find count sketch works well with TPG, even when the sample size is very small.

\section{Theoretical Analysis}
\label{theory}
% theorem of the approximation guarantee
We now analyze  theoretical  properties of the proposed algorithm. We prove that TPG guarantees optimality of the estimated solution, under the assumption that the predictor tensor satisfies Restricted Isometry Property (RIP) \cite{candes2006stable}.  With carefully designed step size,  the algorithm converges to the correct solution in constant number of iterations, and the achievable estimation error depends linearly on the size of the observation error.

We assume the  predictor tensor satisfies  Restricted Isometry Property in the following form:
\begin{definition}{(Restricted Isometry Property)}
	\label{thm:RIP}
	The isometry constant of $\X$ is the smallest number $\delta_R$	such as the following holds for all $\W$ with Tucker rank at most $R$.
	\[(1-\delta_R )  \|\W \|_F^2 \leq   \|   \langle  \X, \W \rangle  \|_F^2 \leq  (1+ \delta_R) \| \W\|_F^2\]
\end{definition}
Note that even though we make the RIP analogy for tensor $\X$, we actually impose the RIP assumption w.r.t. matrix rank instead of tensor rank. Similar assumption can be found in \cite{rauhut2015tensor}.  
%It has been shown that with exponentially high probability, random Gaussian, Bernoulli, and partial Fourier matrices satisfy the RIP condition. 

The proposed solution TPG as in Algorithm  \ref{alg:TPG} is built upon projected gradient method. To prove the convergence, we first guarantee the optimality (local) of the proximal  point step, obtained by ITP in Algorithm \ref{alg:ITP}. The following lemma guarantees the correctness of the solution from ITP.

\begin{lemma}{(Tensor Projection)}
	\label{thm:tensor_projection}
	The projection step  in Algorithm 2, defined as $\T{P}_R:\R^{D_1 \times D_2 \times D_3} \rightarrow  \R^{D_1 \times D_2 \times D_3}$ computes a proximal point $\T{P}_R(  \tilde{\W}^{k+1}) = \W^{k+1}$, whose  Tucker rank is  at most $R$. Formally,
	\begin{eqnarray*}
		\W^{k+1}= \argmin\limits_{\W}\|\W- \tilde{\W}^{k+1}\|_F^2 \quad s.t  \quad {\text{rank}(\W)\leq R} 
	\end{eqnarray*}
  the projected solution $\W^{k+1}$ follows a Tucker model, written as $\W^{k+1} = \T{S} \times_1 \M{U}_1 \times_2 \M{U}_2 \times_3 \M{U}_3$, where each dimension of the core $\T{S}$ is upper bounded by $R$.
\end{lemma}
\proof 
Minimizing $\| \W - \tilde{\W}^{k+1}\|_F^2 $ given $\T{S}$ is equivalent to minimizing the following problem 
\begin{eqnarray*}
	&&\| \T{S} \times_1 \M{U}_1 \times_2 \M{U}_2 \times_3 \M{U}_3 -  \tilde{\W}^{k+1} \|_F^2 \nonumber \\
	&=& \{ \| \T{S} \times_1 \M{U}_1 \times_2 \M{U}_2 \times_3 \M{U}_3 \|_F^2  \\
	&-& 2\langle \T{S} \times_1 \M{U}_1 \times_2 \M{U}_2 \times_3 \M{U}_3 , \tilde{\W}^{k+1} \rangle +   \| \tilde{\W}^{k+1} \|_F^2 \}\\ \nonumber
	&=& \{ \| \T{S} \|_F^2  +   \| \tilde{\W}^{k+1} \|_F^2  \\
	&-& 2\langle \T{S} , \tilde{\W}^{k+1}  \times_1 \M{U}_1^T \times_2 \M{U}_2^T \times_3 \M{U}_3^T \rangle \}\\
	&=& - \|\T{S} \|_F^2 +   \| \tilde{\W}^{k+1} \|_F^2 
\end{eqnarray*}

Given projection matrices $\{\M{U}_n\}$,  we have:
\begin{equation*}
	\T{S} =  \tilde{\W}^{k+1} \times_1 \M{U}_1^T \times_2 \M{U}_2^T \times_3 \M{U}_3^T
\end{equation*}
Thus, the minimizer of $ \|\W- \tilde{\W}^{k+1}\|_F^2$ generates projection matrices that maximize the following  objective  function:
\begin{equation}
 \{\M{U}_n\} = \argmax\limits_{\{\M{U}_n\}} \|\tilde{\W}^{k+1} \times_1 \M{U}_1^T \times_2 \M{U}_2^T \times_3 \M{U}_3^T\|_F^2 
\end{equation}
Each row vector $\V{u}_n$ of $\M{U}_n$ can be solved independently with power iteration. Therefore, repeating this procedure for different modes leads to  an optimal (local) minimizer near a neighborhood of $\tilde{\W}^{k+1}$. In fact, for Tucker tensor, convergence to a saddle point or a local maximum is \textit{only} observed in artificially designed numerical examples \cite{ishteva2011best}. $\blacksquare$

Next we prove our main theorem, which states that  TPG  converges to the correct solution in constant number of iterations with isometry constant $\delta_{2R} < 1/3$.

\begin{theorem}{(Main)}
	\label{thm:main}
%	Bound the error $\|\widehat{\W}- \W^\star\|$:
	For tensor regression model  $\Y = \langle \X, \W \rangle + \E$ , suppose the predictor tensor $\X$ satisfies RIP condition with isometry constant $\delta_{2R} < 1/3 $. Let  $\W^\star$ be the optimal  tensor of Tucker rank at most $R$. Then tensor projected gradient (TPG) algorithm with step-size $ \eta = \frac{1}{1 +\delta_R}  $  computes a feasible solution $\W$ such that  the estimation error $\|\W - \W^\star\|_F^2  \leq    \frac{1}{1-\delta_{2R}}  \|\E \|_F^2 $ in at most $ \lceil  \frac{1}{\log ( 1/\alpha)} \log( \frac{\|\Y\|_F^2}{\|\E|_F^2})\rceil $  iterations for an universal constance $\alpha$ that is independent of problem parameters.
\end{theorem}
\proof : 
The decrease in loss
\begin{small}
	\begin{eqnarray}
		\label{eqn:loss_decrease}
		&&\loss(\W^{k+1})  -  \loss(\W^k) \\
		&=& 	2 \langle  \triangledown \loss (\W^k),  \W^{k+1} -  \W^{k} \rangle + \| \langle  \X, \W^{k+1} -  \W^k\rangle\|_F^2 \nonumber\\
		&\leq& 2 \langle  \triangledown \loss (\W^k),  \W^{k+1} -  \W^{k} \rangle +  (1 +\delta_{2R})  \|\W^{k+1} -  \W^k\ \|_F^2 \nonumber
	\end{eqnarray}
\end{small}
Here the inequality follows from RIP condition. And  isometry constant of $\delta_{2R}$ follows from the subadditivity of rank.
	
Define upper bound 
\begin{small}
	\begin{eqnarray}
	&	u(\W) &:= 2 \langle  \triangledown \loss (\W^k),  \W -  \W^{k} \rangle   +  (1 +\delta_{2R})  \|\W-  \W^k\ \|_F^2  \nonumber \\
	&	=& (1 +\delta_{2R})   \{\|  \W- \tilde{\W}^{k+1} \|^2_F  -  \langle \triangledown \loss (\W^k), \W-\W^k \rangle\} \nonumber
	\end{eqnarray}
\end{small}	
where the second equality follows from the definition of  gradient step $\tilde{\W}^{k+1} =  \W^k - \eta \triangledown \loss (\W^k)$
	
From Equation (\ref{eqn:loss_decrease}) and  Lemma \ref{thm:tensor_projection},
\begin{small}
	\begin{eqnarray}
		\label{eqn:upper_optimal}
		&&\loss(\W^{k+1})  -  \loss(\W^k)  \leq 	u(\W^{k+1}) \leq  	u(\W^\star) \\
		&=&  2\langle  \triangledown \loss (\W^k),  \W^\star -  \W^{k} \rangle   +  (1 +\delta_{2R})  \|\W^\star-  \W^k\ \|_F^2  \nonumber \\ 
		& =&  2\langle  \triangledown \loss (\W^k),  \W^\star -  \W^{k} \rangle + 2\delta_{2R}  \|\W^\star-  \W^k\ \|_F^2  \nonumber \\
		&+& (1 -\delta_{2R})  \|\W^\star-  \W^k\ \|_F^2   \nonumber \\
		&\leq& 2\langle  \triangledown \loss (\W^k),  \W^\star -  \W^{k} \rangle + 2\delta_{2R}  \|\W^\star-  \W^k\ \|_F^2  \nonumber \\
		&+& \| \langle \X,  \W^\star-  \W^k \rangle  \|^2_F  \label{eqn:rip_1}\\
		&=&\loss(\W^{\star})  -  \loss(\W^k)    + 2\delta_{2R}  \|\W^\star-  \W^k\ \|_F^2 \nonumber \\
		&\leq& \loss(\W^{\star})  -  \loss(\W^k) + \tfrac{2\delta_{2R}}{1- \delta_{2R}} \| \langle \X,  \W^\star-  \W^k \rangle  \|^2_F \label{eqn:rip_2}
	\end{eqnarray}
\end{small}
In short, 
\begin{small}
	\begin{eqnarray}
		\label{eqn:recursion }
		\loss(\W^{k+1})   \leq \loss(\W^{\star})  + \frac{2\delta_{2R}  }{1- \delta_{2R}} \| \langle \X,  \W^\star-  \W^k \rangle  \|^2_F 
	\end{eqnarray}
\end{small}
The inequality (\ref{eqn:rip_1}) and (\ref{eqn:rip_2}) follows from RIP condition.
Given model assumption $\Y -\langle \X, \W^\star \rangle  = \E$,   we have 
\begin{small}
	\begin{eqnarray}
		\| \langle \X,  \W^\star-  \W^k \rangle  \|^2_F & = & \| \Y - \langle  \X,\W^k\rangle - \E   \|_F^2 \nonumber \\
		& =&  \loss(\W^k)- 2\langle \E, {\Y -  \langle  \X,\W^k\rangle} \rangle + \| \E\|^2_F \nonumber \\
		&\leq&  \loss(\W^k) +  \frac{2}{C} \loss(\W^k) + \frac{1}{C^2} \loss(\W^k)   \nonumber\\
		&=& (1+\frac{1}{C})^2 \loss (\W^k) 
	\end{eqnarray}
\end{small}
as long as the noise satisfies $C^2\|\E\|_F^2 \leq \loss(\W^k)$.
	
Following Equation (\ref{eqn:recursion }), 
\begin{small}
	\begin{eqnarray}
		\loss(\W^{k+1})  & \leq & \|\E\|_F^2 + \frac{2\delta_{2R}  }{1- \delta_{2R}} (1+\frac{1}{C})^2 \loss (\W^k) \nonumber \\
		& \leq & [\frac{1}{C^2 }+  \frac{2\delta_{2R}  }{1- \delta_{2R}} (1+\frac{1}{C})^2  ] \loss (\W^k) \nonumber\\
		&=& \alpha \loss(\W^k) 
	\end{eqnarray}
\end{small}
With the assumption that $\delta_{2R} < 1/3$, select $C > \tfrac{1+ \delta_{2R}}{1-3\delta_{2R}} $, we have $ [\tfrac{1}{C^2 }+  \tfrac{2\delta_{2R}  }{1- \delta_{2R}} (1+\tfrac{1}{C})^2  ] <1$. The above inequality implies that the algorithm enjoys  a globally geometric convergence rate and the loss decreases multiplicatively.
	
For the simplest case, with initial point as zero, we have  $\loss(\W^0) =  \|\Y\|^2_F $. 
	
In order to obtain a loss value  that is small enough
	\begin{equation}
		\loss(\W^K) \leq  \alpha^K \loss(\W^0)  \leq \| \E\|_F^2
	\end{equation}
	the algorithm requires at least $K \geq  \frac{1}{\log ( 1/\alpha)} \log( \frac{\|\Y\|_F^2}{\|\E|_F^2})$ iterations. 
\begin{small}
	\begin{eqnarray}
		\| \W^K-  \W^\star \|_F^2 &\leq& \frac{1}{1-\delta_{2R}}\|\langle \X, \W^K - \W^\star \rangle\|_F^2 \label{eqn:rip_3} \\
		&\leq&  \frac{1}{1-\delta_{2R}} (1+\frac{1}{C})^2 \loss (\W^K) \nonumber\label{eqn:err_cnt}\\
		&\leq&  \frac{1}{1-\delta_{2R}}   \|\E\|_F^2 \nonumber
	\end{eqnarray}
\end{small}
where inequality  (\ref{eqn:rip_3})  follows from RIP.  $\blacksquare$
	
Theorem \ref{thm:main} shows that under RIP assumption, TGP converges to an approximate solution in $O(\tfrac{1}{\log ( 1/\alpha)} \log( \tfrac{\|\Y\|_F^2}{\|\E|_F^2}) )$  number of iterations, which depends solely on the logarithm of  signal to noise ratio. 
The achievable  estimation error depends linearly on the size of the observation error. 

As a pre-processing step, our proposed algorithm employs $l_2$-subspace embedding (count sketch) to subsample the data instances.  
\begin{definition}{($l_2$-subspace embedding)}
	A $(1 \pm \epsilon)$ $l_2$-subspace embedding for the column space of a $N \times D$ matrix $\M{A}$ is a matrix $\M{S}$ for which for all $\V{x} \in \R^D$
	\[ \| \M{S} \M{A} \V{x} \|_2^2   = ( 1  \pm \epsilon) \|\M{A}\V{x}\|_2^2\]
\end{definition}

Subsampling step solves an unconstrained optimization problem which  is essentially a least square problem, the approximation error  can be upper-bounded as follows:
\begin{lemma}{(Approximation Guarantee)}
	For any $0<\delta <1$, given a sparse $l_2$-subspace embedding matrix $\M{S}$  with $K = O((D_1D_2D_3)^2/ \delta \epsilon^2)$ rows, then with probability $(1-\delta)$,  we can achieve $(1+\epsilon)$-approximation. The sketch $\X\times_1 \M{S}$ can be computed in $O(nnz(\X))$ time, and  $\Y \times_1 \M{S}$ can be computed in $O(nnz(\Y))$ time.
\end{lemma}

The result follows directly from \cite{clarkson2013low}. The randomized sketching leads to a $(1+\epsilon)$ approximation of  the original tensor regression solution. It also serves as a noise reduction step, facilitating fast convergence of subsequent TPG procedure.

\section{Applications of Tensor Regression}
\label{application}
Tensor regression finds applications in many domains. We present two examples:  one is the multi-linear multi-task learning problem in machine learning community, and the other is the spatio-temporal forecasting problem in time series analysis domain.

\subsection{Multi-linear Multi-task Learning}
Multi-linear multi-task learning \cite{romera2013multilinear, wimalawarne2014multitask}  tackles the scenario  where the tasks to be learned are references by multiple indices, thus contain multi-modal relationship. Given the predictor and response for each task: $(\M{X}^t \in \R^{m_t \times d_t}, \M{Y}^t \in \R^{m_t \times 1})$, traditional multi-task learning concatenate parameter vector $\V{w}^t \in \R^{d_t \times 1}$ into a matrix. Here, with additional information about task indices, the model stacks the coefficient vectors into a model tensor $\W$.  The empirical loss is defined as the summarization of the least square loss for all the tasks, i.e $\loss(\W; \X, \Y) = \sum_{t=1}^T \| \M{Y}^t -  \M{X}^t \V{w }^t \|^2_F$, with $\V{w}^t$ as the $t$ th column of $\W_{(1)}$.   The multi-linear multi-task learning problem can be described as follows: 
\begin{eqnarray}\label{eqn:cokriging}
\widehat{\W} = \argmin\limits_{\W} \left\{ \sum_{t=1}^T \| \M{Y}^t -  \M{X}^t \V{w }^t \|^2_F \right \}  \nonumber \\
 \text{s.t.} \quad  \text{rank} (\W) \leq R 
\end{eqnarray}
%\subsection{Multimodel Ensemble}
%The multi-model ensemble \cite{tebaldi2007use} problem arises in climate modeling.  It aims to combine climate model outputs into a more accurate description of the observations.  Suppose we have gathered the model simulation outputs from $S$ models of $M$ climate variables in $P$ locations over time period $T$ as  $\Y \in \R^{P\times T \times M \times S}$. At the same time, we are given access to the actual observations of the same variables, locations and time  $\X \in \R^{P\times T\times M}$.  We build a linear model  as $\X = \mathcal{W}_{:,:, m}\mathbf{Y}_{t, m} $, where $\mathbf{Y}_{t, m} = [\Y_{:, t, m,1}^{\top}, \ldots, \Y_{:, t, m, S}^{\top}]^{\top}$ denotes the concatenation of $S$ model outputs at time $t$ for variable $m$, and  $\W\in \R^{P \times PS \times M}$ characterizes the ``importance'' of various models in climate predictions. We formulate the multi-model ensemble task as follows: 
%\begin{eqnarray}\label{eqn:multimodel}
%&\widehat{\W} = \argmin\limits_{\W} \left\{ \| \widehat{\X}- \X \|^2_F +  \mu \sum\limits_{m=1}^M \text{tr} (\widehat{\X}_{:,:,m}^\top \mathbf{L} \widehat{\X}_{:,:,m}) \right\} \nonumber\\
%&\text{s.t.}\  \widehat{\X}_{:,t,m} = \mathcal{W}_{:,:, m}\mathbf{Y}_{t, m}, \ \sum\limits_{n=1}^N \text{rank}(\W_{(n)}) \leq R \nonumber
%\end{eqnarray}

\subsection{Spatio-temporal Forecasting}
Spatio-temporal forecasting  \cite{cressie2015statistics}  is to predict the future values  given their historical measurements. Suppose we are given access to measurements  $\X \in \mathcal{R}^{T\times P \times M}$ of $T$ timestamps of $M$  variables over $P$ locations as well as  the geographical coordinates of $P$ locations.  We can  model the time series with a Vector Auto-regressive (VAR) model  of lag $L$, where we assume the generative process as
$\X_{t,:,m} = \mathbf{X}_{t, m} \mathcal{W}_{:,:, m}+ \mathcal{E}_{t,:,m}, \text{for } m = 1, \ldots, M \text{ and } t = L+1, \ldots, T$. Here  $\mathbf{X}_{t, m} = [\X_{ t-1,:, m}^{\top}, \ldots, \X_{t-L,:,  m}^{\top}]$ denotes the concatenation of $L$-lag historical data before time $t$. 
We learn a model coefficient tensor $\W \in \R^{PL \times P \times M}$ to forecast multiple variables simultaneously. The forecasting task can be formulated as follows:
\begin{eqnarray}\label{eqn:forecasting}
	& \widehat{\W} = \argmin\limits_{\W} \left\{ \| \widehat{\X}- \X \|^2_F +  \mu \sum\limits_{m=1}^M \text{tr} (\widehat{\X}_{:,:,m}^\top \mathbf{L} \widehat{\X}_{:,:,m}) \right\}  \nonumber \\
	& \text{s.t.} \ \widehat{\X} = \mathbf{X}_{t, m} \mathcal{W}_{:,:, m},  \  \text{s.t.} \quad  \text{rank} (\W) \leq R  
\end{eqnarray}
where rank constraint imposes structures such as spatial clustering and temporal periodicity on the model.  The Laplacian regularizer $\mathbf{L}$ is constructed from the kernel using the geographical information, which accounts for the spatial proximity of observations.

%Typical setting is passive recovery

\section{Experiments}
\label{exp}
%experiment
\begin{figure*}[h]
	%\vskip 0.2in
	\centering 
	\subfigure[RMSE]{
		\includegraphics[scale = 0.15]{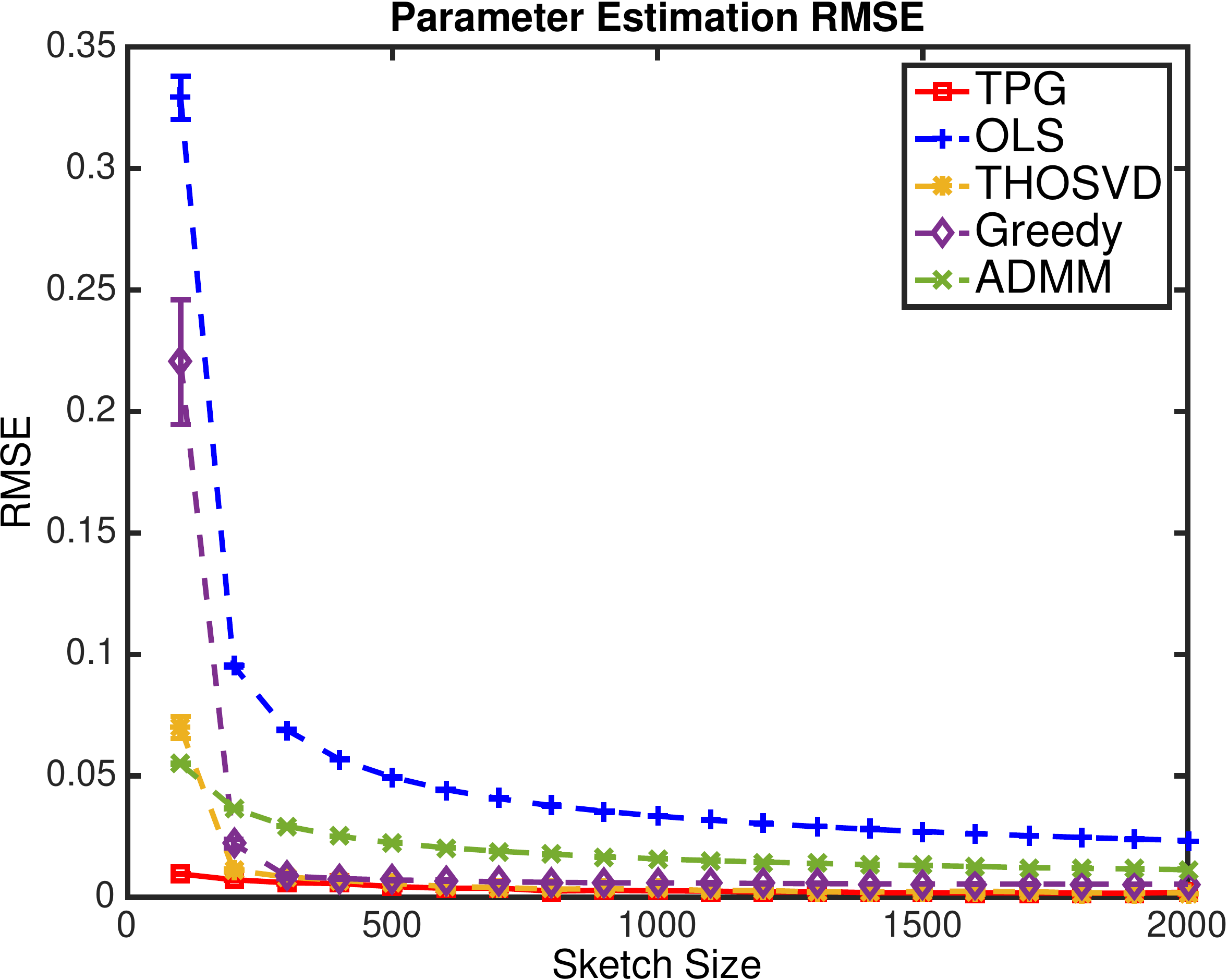}
		\label{fig:synth_err}
	}
	\subfigure[Run Time]{
		\includegraphics[scale = 0.15]{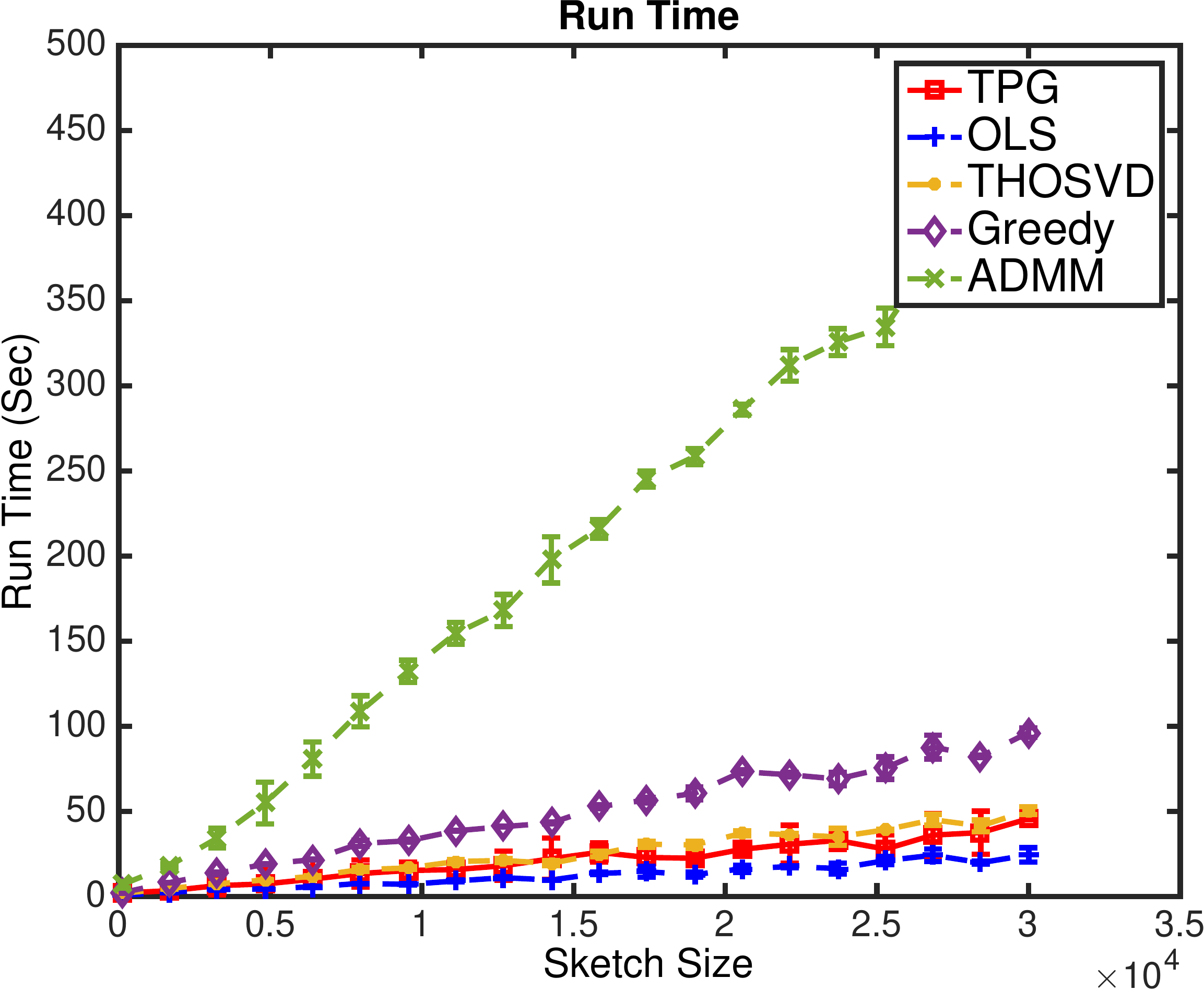}
		\label{fig:synth_runtime}
	}
	\subfigure[RMSE]{
		\includegraphics[scale = 0.15]{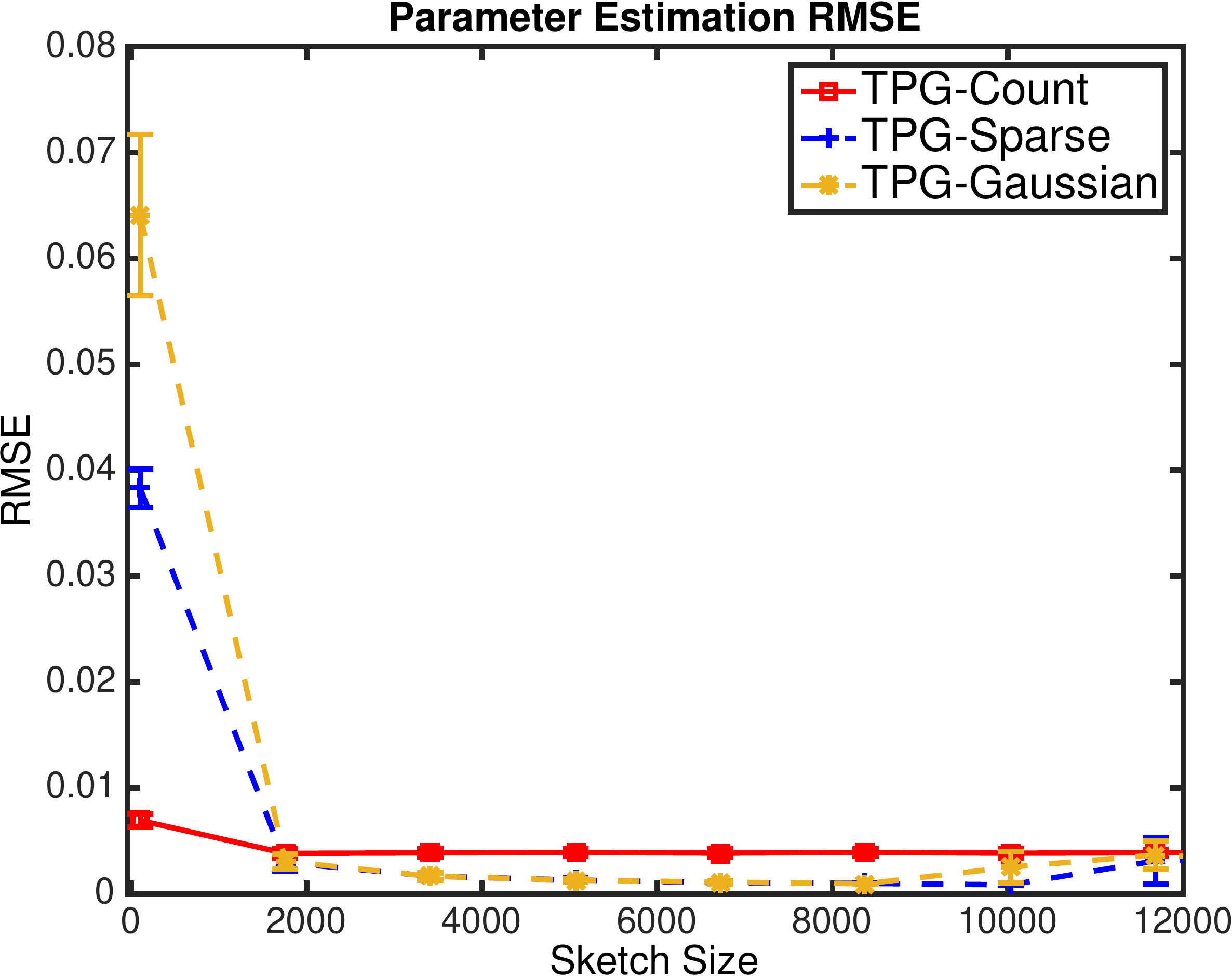}
		\label{fig:synth_skt_err}
	}
	\subfigure[Run Time]{
		\includegraphics[scale = 0.15]{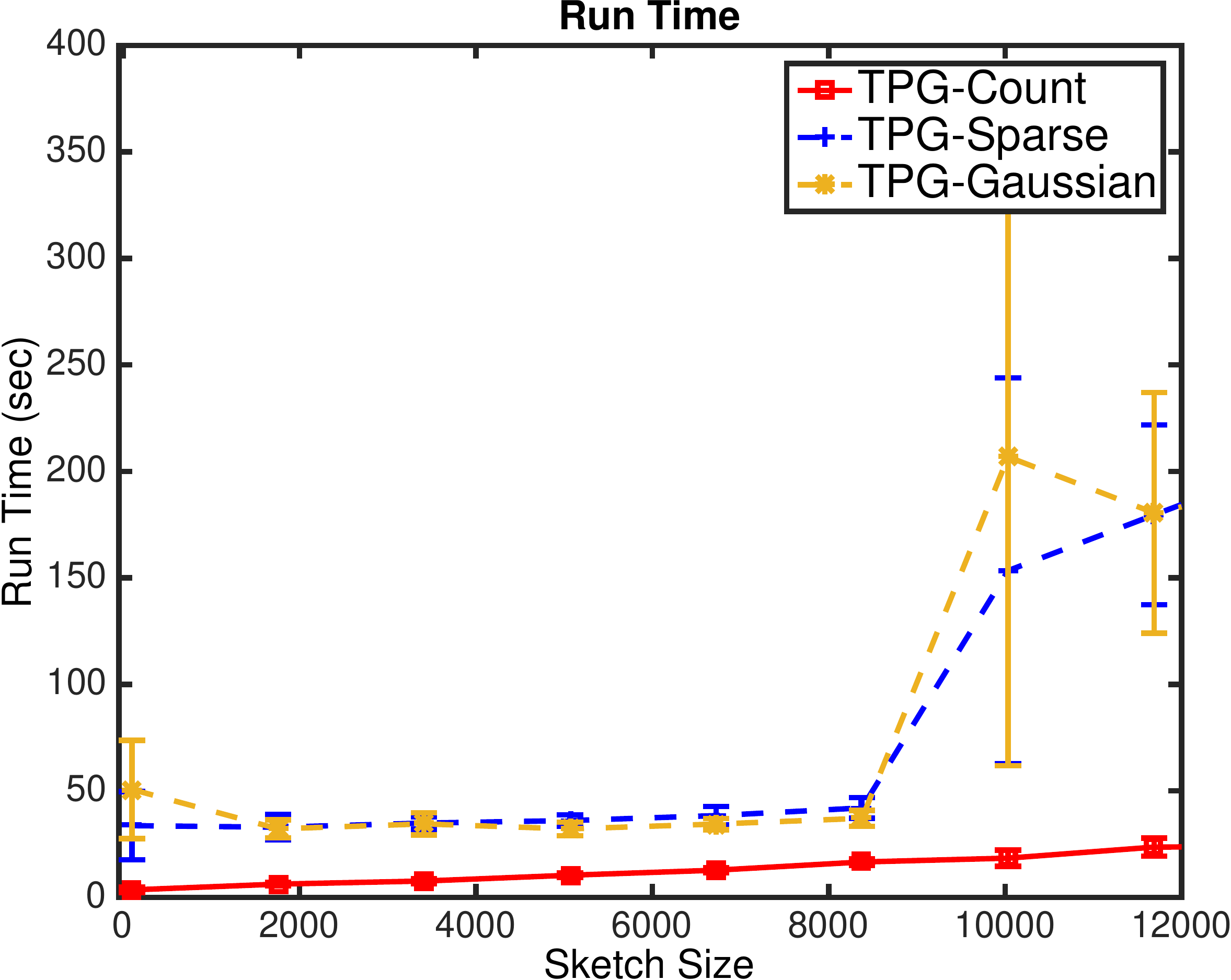}
		\label{fig:synth_skt_runtime}
	}

	\caption{Performance comparison on the synthetic dataset over 10 random runs.  \subref{fig:synth_err}  model parameter estimation RMSE for different algorithms,  \subref{fig:synth_runtime} running time  with respect to sketching size  for different algorithms,  \subref{fig:synth_skt_err}  RMSE for different sketching method,  \subref{fig:synth_skt_runtime} running time for different sketching method.}
	\vskip -0.2in
	\label{fig:synthetic}
\end{figure*} 

We  conduct a set of experiments on one synthetic dataset and two real world applications. In this
section, we present and analyze the results obtained. We compare TGP with following baseline methods:
\begin{itemize}
\item OLS: ordinary least square estimator without low-rank constraint
\item THOSVD  \cite{de2000multilinear}: a two-step heuristic that first solves the least square and then performs truncated singular value decomposition
\item Greedy \cite{yu2014fast}: a fast tensor regression solution that sequentially estimates rank one subspace based on Orthogonal Matching Pursuit   
\item ADMM \cite{gandy2011tensor}: alternating direction method of multipliers for nuclear norm regularized optimization 
\end{itemize}

\subsection{Synthetic Dataset }
We construct a model coefficient tensor $\W$ of size $30 \times 30 \times 20 $ with Tucker rank equals to $2$ for all modes.  Then, we generate the observations $\Y$ and $\X$ according to multivariate regression model $\Y_{:,:,m} =\X_{:, :, m} \W_{:,:,m} + \mathcal{E}_{:,:,m}$ for $m = 1, \ldots, M$, where $\mathcal{E}$ is a noise tensor with zero mean Gaussian  elements. We generate the time series with  $30,000$ time stamps and repeat the procedure for $10$ times. 

Figure \ref{fig:synth_err} and \ref{fig:synth_runtime} shows the  parameter estimation RMSE and the run time error bar with respect to the sketching size.  Since the true model is low-rank, simple OLS suffers from poor performance. Other methods are able to converge to the correct solution. The main difference occurs for  small sketch size scenario.  TPG demonstrates its impressively robustness to noise while others shows high variance in estimation RMSE. ADMM obtains reasonable accuracy and is relatively robust, but is very slow.

We also investigate the impact of  sketching scheme on TPG. We  compare  count sketch (Count) with sketch with i.i.d Gaussian entries (Gaussian) and sparse random projection (Sparse) \cite{achlioptas2003database}.  Figure  \ref{fig:synth_skt_err} and \ref{fig:synth_skt_runtime} shows the  parameter estimation RMSE and the run time errorbar  for TPG combined with different random sketching algorithm.  TPG with count sketch achieves best performance, especially for small sketch size. The results justify the metrit of leveraging count sketch for noise reduction, in order to accelerate the convergence of TPG algorithm.

\subsection{Real Data}
In this section, we test the described approaches with two real world application datasets: multi-linear multi-task learning and  spatio-temporal forecasting problem. 

\subsubsection{Multi-linear Multi-task Learning}
We compare TPG with state-of-art multi-linear multi-task baseline.  Our evaluation follows the same experiment setting in \cite{romera2013multilinear} on the restaurant \& consumer dataset, provided by the  authors in the paper. The  data set contains consumer ratings given to different restaurants. The data has $138$ consumers gave $3$ type of scores for restaurant service. Each restaurant has $45$ attributes for rating prediction. The total number of instances for all the tasks were $3483$. The problem results in a model tensor of dimension $45 \times 138 \times 3$.

We split the training and testing set with different ratio ranging from 0.1 to 0.9 and randomly select the training data instances. When the training size is small, many tasks contained no training example. We also select 200 instances as the validation set. We compare with MLMTL-C and multi-task feature learning baselines in  the original paper. MTL-L21 uses $L_{21}$ regularizer and MTL-Trace is the trace-norm regularized multi-task learning algorithm. The model parameters  are selected by minimizing the mean squared error on the validation set.
\begin{figure}[t]
	%\vskip 0.2in
	\centering 
	\subfigure[MSE]{
		\includegraphics[scale = 0.15]{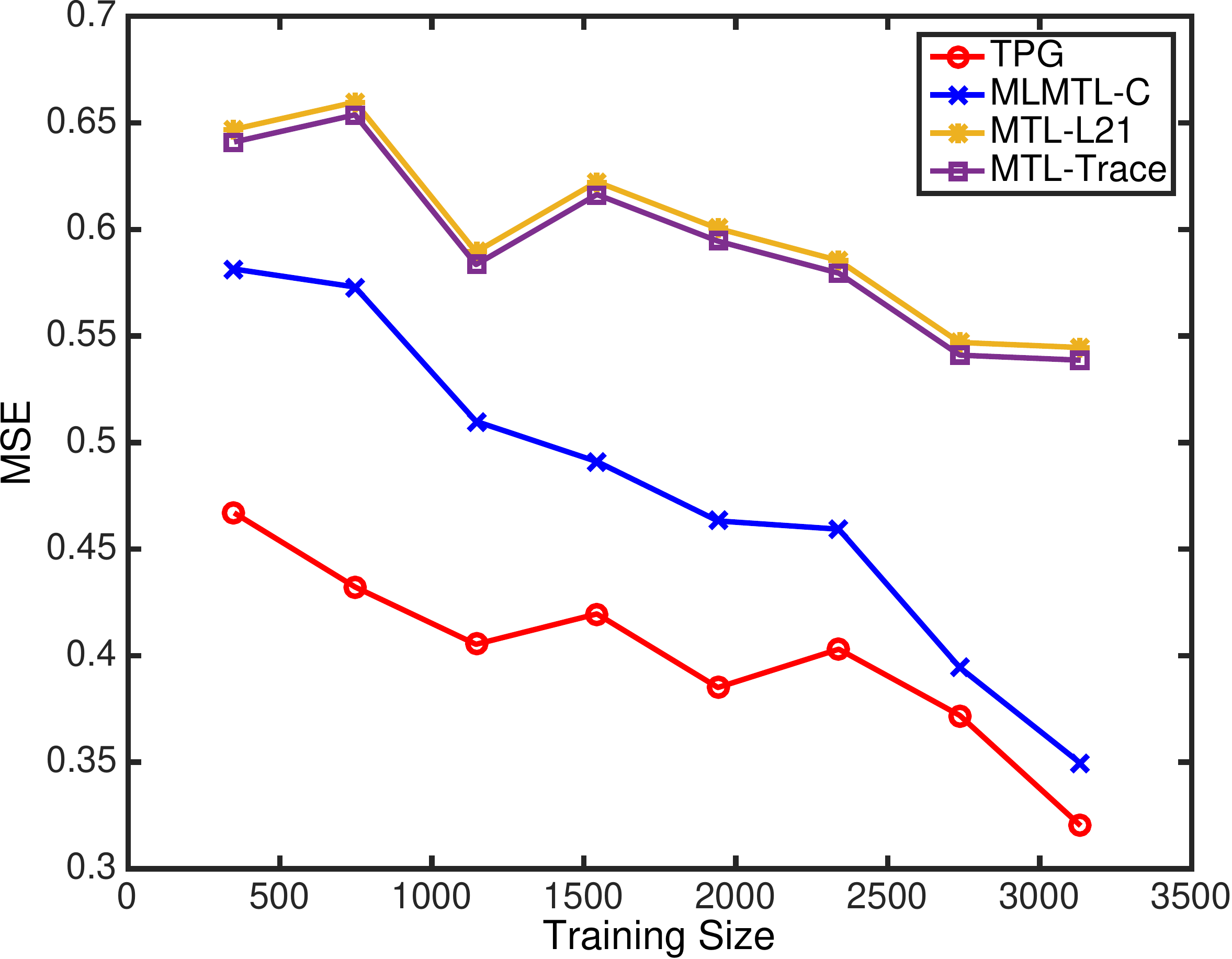}
		\label{fig:mlmtl_err}
	}
	\subfigure[Run Time]{
		\includegraphics[scale = 0.15]{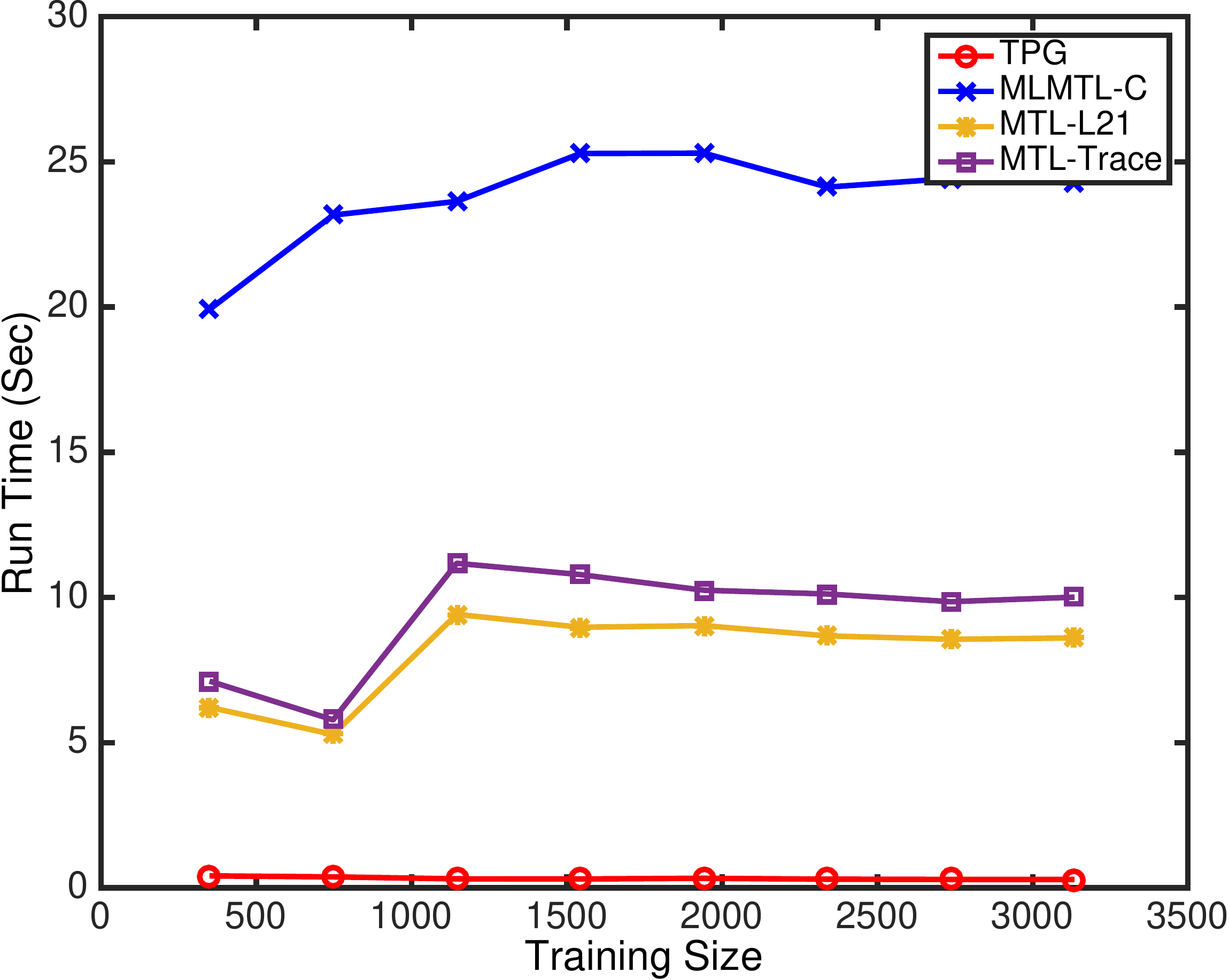}
		\label{fig:mlmtl_runtime}
	}
	\caption{Multi-linear multi-task learning  performance comparison on the synthetic dataset over 10 random runs.  \subref{fig:mlmtl_err} average forecasting MSE and  \subref{fig:mlmtl_runtime} running time w.r.t. training size.}
	\vskip -0.2in
	\label{fig:mlmtl}
\end{figure} 

 Figure \ref{fig:mlmtl} demonstrates the prediction performance in terms of MSE and runtime with respect to different training size. Compared with MLMTL-C, TPG is  around $10 \% - 25\% $ more accurate  and at least $20$ times faster. MTL-L21 or MTL-Trace runs faster than MLMTL-C but also sacrifices accuracy. The difference is more noticeable in the case of  small training size.  These results are not surprising. Given limited samples and highly correlated tasks in the restaurant score prediction,  the model parameters demonstrate low-rank property. In fact, we found that rank $1$ is the optimal setting for this data during the experiments.

\subsubsection{Spatio-Temporal Forecasting}
For the spatio-temporal forecasting task, we experiment with following two datasets.
\paragraph{Foursquare Check-In}
The Foursquare check-in data set contains the users’ check-in records in Pittsburgh area
from Feb 24 to May 23, 2012, categorized by different venue types such as Art \& Entertainment,
College \& University, and Food.  We extract 
hourly check-in records of $739$  users in $34$ different  venues categories over $3,474$ hours time period as well as users' friendship network.
\paragraph{USHCN Measurements}
The U.S. Historical Climatology Network  (USHCN) daily  (\url{http://cdiac.ornl.gov/ftp/ushcn\_daily/}) contains  daily measurements for $5$ climate variables for more than $100$ years. The five climate variables correspond to  temperature max, temperature min, precipitation, snow fall and snow depth. The records were collected across more than $1,200$ locations and spans over  $45,384$ time stamps.

 We split the data along the temporal dimension into 80\% training set and 20\% testing set. We choose VAR (3) model and use 5-fold cross-validation to select the rank during the training phase.  For both datasets,  we normalize each individual time series by removing the mean and dividing by standard deviation.  Due to the memory constraint of the Greedy algorithm,  evaluations are conducted on down-sampled datasets.
  \begin{table}[h]
  			\vspace{-2mm}
 	\caption{ Forecasting RMSE and run time on Foursquare check-in data  and USHCN daily measurement for VAR process with 3 lags, trained with 80\%  of the time series.} %title of the table
 	\small
 	\label{tab:real_RMSE}
 	%\vskip 0.15in
 	\begin{center}
 		\begin{tiny}
 			\begin{sc}
 				\centering  \footnotesize% centering table
 				\begin{tabular}{@{}c@{\;\;} c @{\;\;} c @{\;\;}c @{\;\;} c@{\;\;} c @{\;\;}c@{\;\;} } % creating eight columns
 					\hline
 					\hline
 					%\abovespace\belowspace
 					& TPG  & OLS & THOSVD   &  Greedy  & ADMM     \\
 					\hline
 					RMSE  & \textbf{0.3580} & 0.8277 &0.4780 &0.3639 & 0.3916 \\
 					RunTime &37.06 &5.85 &	12.37&	290.12	&445.41 \\
 					%Climate 4  & & & 0.9511 & 1.1374 & 1.1374 & 0.9449  & 0.9342 & 1.1255\\
 					\hline
 				\end{tabular}
 			\end{sc}
 			\begin{sc}
 				\centering  \footnotesize% centering table
 				\begin{tabular}{@{}c@{\;\;} c @{\;\;} c @{\;\;}c @{\;\;} c@{\;\;} c @{\;\;}c@{\;\;} } % creating eight columns
 					\hline
 					\hline
 					%\abovespace\belowspace
 					& TPG  & OLS & THOSVD   &  Greedy  & ADMM     \\
 					\hline
 					RMSE  &\textbf{0.3872}& 1.4265 & 0.7224 & 	0.4389 &  0.5893\\
 					RunTime & 144.43 &23.69 &	46.26 &	410.38	& 6786\\
 					%Climate 4  & & & 0.9511 & 1.1374 & 1.1374 & 0.9449  & 0.9342 & 1.1255\\
 					\hline
 				\end{tabular}
 			\end{sc}
 		\end{tiny}
 	\end{center}
 	\vspace{-0.3in}
 \end{table}
 
Table \ref{tab:real_RMSE} presents the best forecasting performance (w.r.t sketching size) and the corresponding run time for each of the methods. TPG outperforms baseline methods with higher accuracy.  Greedy shows similar accuracy, but TPG converges in very few iterations. For USHCN, TPG achieves much higher accuracy with significantly shorter run time.  Those results demonstrate the efficiency of our proposed algorithm for spatio-temporal forecasting tasks.

\begin{figure}[h]
	%\vskip 0.2in
	\centering
		\includegraphics[scale = 0.16]{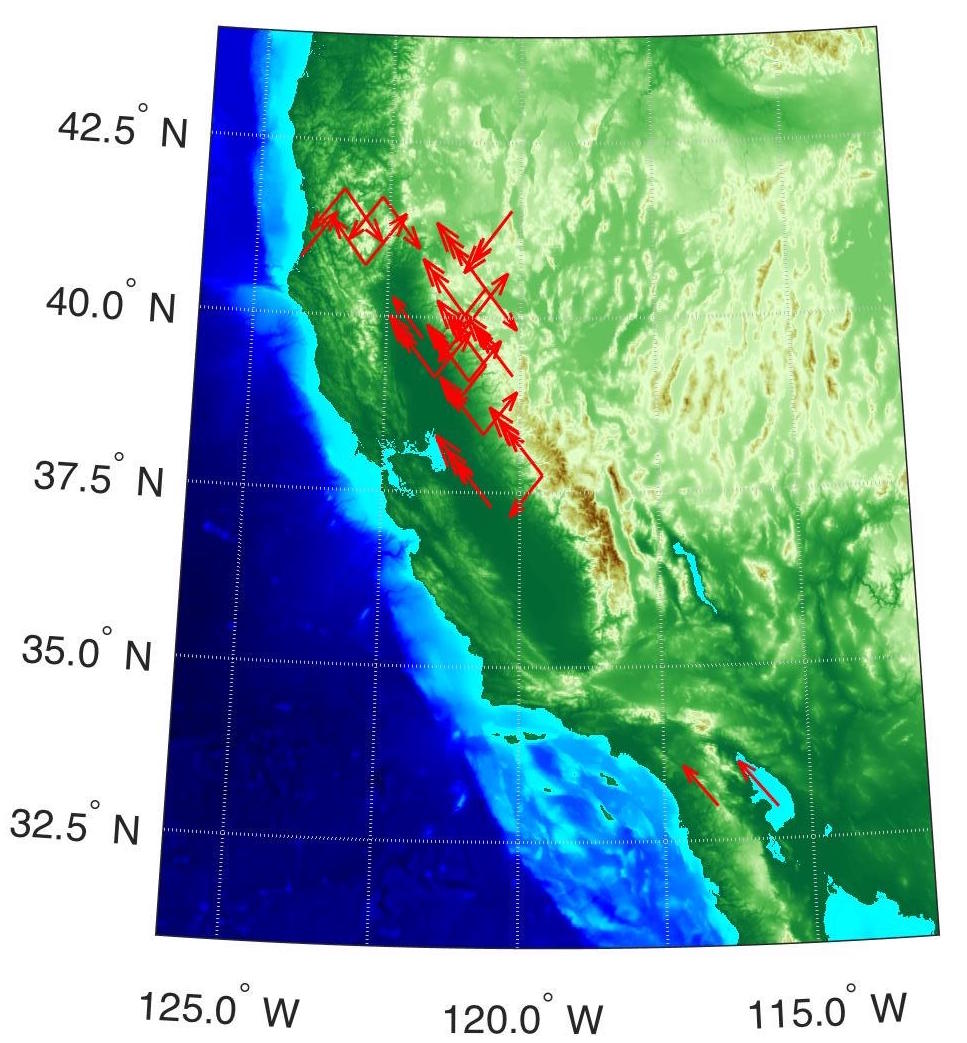}
		\vspace{-2mm}
	\caption{Velocity vectors plot of spatial-temporal dependency graph obtained via TPG.  Results are averaged across all five different climate variables.}
	\label{fig:arrow_map}
	 	\vspace{-0.3in}
\end{figure}

We further investigate the learned structure of TPG algorithm from USHCN data.  Figure \ref{fig:arrow_map} shows the spatial-temporal dependency graph on the terrain of California. Each velocity vector represents the aggregated weight learned by TPG from one location to the other.  The graph provides an interesting illustration of atmospheric circulation. For example, near Shasta-Trinity National forecast in northern California, the air flow into the forecasts. On the east side along Rocky mountain area,  there is a strong atmospheric pressure, leading to wind moving from south east to north west passing the bay area. Another notable atmospheric circulation happens near Salton sea at the border of Utah, caused mainly by the evaporation of the sea.

\section{Discussion}
\label{disc}
The implication of our approach has several interesting aspects that might shed light upon future algorithmic design.

(1) The  projection step  in TPG does not depends on data, thus it connects to tensor decomposition techniques such as high order orthogonal iteration (HOOI)  \cite{de2000best}. However, there is subtle difference in that the regression  would call for early stop of iterative projection as it sequentially search for the orthogonal subspaces.

(2) TPG behaves similarly as first order methods. The convergence rate can be further improved with second order Newton method. This can be done easily by replacing the gradient with Hessian.  This modification does not affect the theoretical properties of the proposed algorithm, but would lead to significant empirical improvement \cite{jain2010guaranteed}.

%(3) The projection step only guarantees the local optimality of ITP procedure. We bound the estimation error based on the fact that  stationary points are close to each other, and the choice of local optima does not differ significantly. More rigorous analysis can be derived by analyzing the  distribution of  singular values for random tensors.

\section{Conclusion}
\label{con}
% conclustion
 In this paper, we study tensor regression as a tool to analyze multiway data. We introduce Tensor Projected Gradient to solve the problem. Our approach is built upon projected gradient method, generalizing iterative hard thresholding technique to high order tensors.  The algorithm is very simple and general, which can be  easily applied to many tensor regression models.  It also shares the efficiency of iterative hard thresholding method. We prove that the algorithm converges within a constant number of iterations and the achievable estimation error is linear to the size of the noise. We  evaluate our method on multi-linear multi-task learning as well as  spatio-temporal forecasting applications. The results show that the our  method  is significantly faster and is impressively robust to noise.

\section{Acknowledgment}
This work is supported in part by the U. S. Army Research
Office under grant number W911NF-15-1-0491, NSF Research Grant IIS-
1254206 and IIS-1134990.
The views and conclusions are those of the authors and
should not be interpreted as representing the official policies
of the funding agency, or the U.S. Government.

{\small
\bibliography{icml2016}

\begin{thebibliography}{28}
\providecommand{\natexlab}[1]{#1}
\providecommand{\url}[1]{\texttt{#1}}
\expandafter\ifx\csname urlstyle\endcsname\relax
  \providecommand{\doi}[1]{doi: #1}\else
  \providecommand{\doi}{doi: \begingroup \urlstyle{rm}\Url}\fi

\bibitem[Acar et~al.(2007)Acar, Aykut-Bingol, Bingol, Bro, and
  Yener]{acar2007multiway}
Acar, Evrim, Aykut-Bingol, Canan, Bingol, Haluk, Bro, Rasmus, and Yener,
  B{\"u}lent.
\newblock Multiway analysis of epilepsy tensors.
\newblock \emph{Bioinformatics}, 23\penalty0 (13):\penalty0 i10--i18, 2007.

\bibitem[Achlioptas(2003)]{achlioptas2003database}
Achlioptas, Dimitris.
\newblock Database-friendly random projections: Johnson-lindenstrauss with
  binary coins.
\newblock \emph{Journal of computer and System Sciences}, 66\penalty0
  (4):\penalty0 671--687, 2003.

\bibitem[Blumensath \& Davies(2009)Blumensath and
  Davies]{blumensath2009iterative}
Blumensath, Thomas and Davies, Mike~E.
\newblock Iterative hard thresholding for compressed sensing.
\newblock \emph{Applied and Computational Harmonic Analysis}, 27\penalty0
  (3):\penalty0 265--274, 2009.

\bibitem[Calamai \& Mor{\'e}(1987)Calamai and Mor{\'e}]{calamai1987projected}
Calamai, Paul~H and Mor{\'e}, Jorge~J.
\newblock Projected gradient methods for linearly constrained problems.
\newblock \emph{Mathematical programming}, 39\penalty0 (1):\penalty0 93--116,
  1987.

\bibitem[Candes et~al.(2006)Candes, Romberg, and Tao]{candes2006stable}
Candes, Emmanuel~J, Romberg, Justin~K, and Tao, Terence.
\newblock Stable signal recovery from incomplete and inaccurate measurements.
\newblock \emph{Communications on pure and applied mathematics}, 59\penalty0
  (8):\penalty0 1207--1223, 2006.

\bibitem[Cichocki et~al.(2009)Cichocki, Zdunek, Phan, and
  Amari]{cichocki2009nonnegative}
Cichocki, Andrzej, Zdunek, Rafal, Phan, Anh~Huy, and Amari, Shun-ichi.
\newblock \emph{Nonnegative matrix and tensor factorizations: applications to
  exploratory multi-way data analysis and blind source separation}.
\newblock John Wiley \& Sons, 2009.

\bibitem[Clarkson \& Woodruff(2013)Clarkson and Woodruff]{clarkson2013low}
Clarkson, Kenneth~L and Woodruff, David~P.
\newblock Low rank approximation and regression in input sparsity time.
\newblock In \emph{Proceedings of the forty-fifth annual ACM symposium on
  Theory of computing}, pp.\  81--90. ACM, 2013.

\bibitem[Cressie \& Wikle(2015)Cressie and Wikle]{cressie2015statistics}
Cressie, Noel and Wikle, Christopher~K.
\newblock \emph{Statistics for spatio-temporal data}.
\newblock John Wiley \& Sons, 2015.

\bibitem[De~Lathauwer et~al.(2000{\natexlab{a}})De~Lathauwer, De~Moor, and
  Vandewalle]{de2000best}
De~Lathauwer, Lieven, De~Moor, Bart, and Vandewalle, Joos.
\newblock On the best rank-1 and rank-(r 1, r 2,..., rn) approximation of
  higher-order tensors.
\newblock \emph{SIAM Journal on Matrix Analysis and Applications}, 21\penalty0
  (4):\penalty0 1324--1342, 2000{\natexlab{a}}.

\bibitem[De~Lathauwer et~al.(2000{\natexlab{b}})De~Lathauwer, De~Moor, and
  Vandewalle]{de2000multilinear}
De~Lathauwer, Lieven, De~Moor, Bart, and Vandewalle, Joos.
\newblock A multilinear singular value decomposition.
\newblock \emph{SIAM journal on Matrix Analysis and Applications}, 21\penalty0
  (4):\penalty0 1253--1278, 2000{\natexlab{b}}.

\bibitem[Eckart \& Young(1936)Eckart and Young]{eckart1936approximation}
Eckart, Carl and Young, Gale.
\newblock The approximation of one matrix by another of lower rank.
\newblock \emph{Psychometrika}, 1\penalty0 (3):\penalty0 211--218, 1936.

\bibitem[Gandy et~al.(2011)Gandy, Recht, and Yamada]{gandy2011tensor}
Gandy, Silvia, Recht, Benjamin, and Yamada, Isao.
\newblock Tensor completion and low-n-rank tensor recovery via convex
  optimization.
\newblock \emph{Inverse Problems}, 27\penalty0 (2):\penalty0 025010, 2011.

\bibitem[Garg \& Khandekar(2009)Garg and Khandekar]{garg2009gradient}
Garg, Rahul and Khandekar, Rohit.
\newblock Gradient descent with sparsification: an iterative algorithm for
  sparse recovery with restricted isometry property.
\newblock In \emph{Proceedings of the 26th Annual International Conference on
  Machine Learning}, pp.\  337--344. ACM, 2009.

\bibitem[Hasselmann(1997)]{hasselmann1997multi}
Hasselmann, Klaus.
\newblock Multi-pattern fingerprint method for detection and attribution of
  climate change.
\newblock \emph{Climate Dynamics}, 13\penalty0 (9):\penalty0 601--611, 1997.

\bibitem[Ishteva et~al.(2011{\natexlab{a}})Ishteva, Absil, Van~Huffel, and
  De~Lathauwer]{ishteva2011best}
Ishteva, Mariya, Absil, P-A, Van~Huffel, Sabine, and De~Lathauwer, Lieven.
\newblock Best low multilinear rank approximation of higher-order tensors,
  based on the riemannian trust-region scheme.
\newblock \emph{SIAM Journal on Matrix Analysis and Applications}, 32\penalty0
  (1):\penalty0 115--135, 2011{\natexlab{a}}.

\bibitem[Ishteva et~al.(2011{\natexlab{b}})Ishteva, Absil, Van~Huffel, and
  De~Lathauwer]{ishteva2011tucker}
Ishteva, Mariya, Absil, P-A, Van~Huffel, Sabine, and De~Lathauwer, Lieven.
\newblock Tucker compression and local optima.
\newblock \emph{Chemometrics and Intelligent Laboratory Systems}, 106\penalty0
  (1):\penalty0 57--64, 2011{\natexlab{b}}.

\bibitem[Jain et~al.(2010)Jain, Meka, and Dhillon]{jain2010guaranteed}
Jain, Prateek, Meka, Raghu, and Dhillon, Inderjit~S.
\newblock Guaranteed rank minimization via singular value projection.
\newblock In \emph{Advances in Neural Information Processing Systems}, pp.\
  937--945, 2010.

\bibitem[Kolda \& Bader(2009)Kolda and Bader]{kolda2009tensor}
Kolda, Tamara~G and Bader, Brett~W.
\newblock Tensor decompositions and applications.
\newblock \emph{SIAM review}, 51\penalty0 (3):\penalty0 455--500, 2009.

\bibitem[Rauhut et~al.(2015)Rauhut, Schneider, and Stojanac]{rauhut2015tensor}
Rauhut, Holger, Schneider, Reinhold, and Stojanac, {\v{Z}}eljka.
\newblock Tensor completion in hierarchical tensor representations.
\newblock In \emph{Compressed Sensing and its Applications}, pp.\  419--450.
  Springer, 2015.

\bibitem[Rockafellar(1976)]{rockafellar1976monotone}
Rockafellar, R~Tyrrell.
\newblock Monotone operators and the proximal point algorithm.
\newblock \emph{SIAM journal on control and optimization}, 14\penalty0
  (5):\penalty0 877--898, 1976.

\bibitem[Romera-Paredes et~al.(2013)Romera-Paredes, Aung, Bianchi-Berthouze,
  and Pontil]{romera2013multilinear}
Romera-Paredes, Bernardino, Aung, Hane, Bianchi-Berthouze, Nadia, and Pontil,
  Massimiliano.
\newblock Multilinear multitask learning.
\newblock In \emph{Proceedings of the 30th International Conference on Machine
  Learning}, pp.\  1444--1452, 2013.

\bibitem[Signoretto et~al.(2014)Signoretto, Dinh, De~Lathauwer, and
  Suykens]{signoretto2014learning}
Signoretto, Marco, Dinh, Quoc~Tran, De~Lathauwer, Lieven, and Suykens,
  Johan~AK.
\newblock Learning with tensors: a framework based on convex optimization and
  spectral regularization.
\newblock \emph{Machine Learning}, 94\penalty0 (3):\penalty0 303--351, 2014.

\bibitem[Vasilescu \& Terzopoulos(2002)Vasilescu and
  Terzopoulos]{vasilescu2002multilinear}
Vasilescu, M Alex~O and Terzopoulos, Demetri.
\newblock Multilinear analysis of image ensembles: Tensorfaces.
\newblock In \emph{Computer Vision—ECCV 2002}, pp.\  447--460. Springer,
  2002.

\bibitem[Wimalawarne et~al.(2014)Wimalawarne, Sugiyama, and
  Tomioka]{wimalawarne2014multitask}
Wimalawarne, Kishan, Sugiyama, Masashi, and Tomioka, Ryota.
\newblock Multitask learning meets tensor factorization: task imputation via
  convex optimization.
\newblock In \emph{Advances in Neural Information Processing Systems}, pp.\
  2825--2833, 2014.

\bibitem[Yu et~al.(2014)Yu, Bahadori, and Liu]{yu2014fast}
Yu, Rose, Bahadori, Mohammad~Taha, and Liu, Yan.
\newblock Fast multivariate spatio-temporal analysis via low rank tensor
  learning.
\newblock \emph{NIPS}, 2014.

\bibitem[Yu et~al.(2015)Yu, Cheng, and Liu]{yu2015accelerated}
Yu, Rose, Cheng, Dehua, and Liu, Yan.
\newblock Accelerated online low rank tensor learning for multivariate
  spatiotemporal streams.
\newblock In \emph{Proceedings of the 32nd International Conference on Machine
  Learning (ICML-15)}, pp.\  238--247, 2015.

\bibitem[Zhao et~al.(2011)Zhao, Caiafa, Mandic, Zhang, Ball, Schulze-Bonhage,
  and Cichocki]{zhao2011multilinear}
Zhao, Qibin, Caiafa, Cesar~F, Mandic, Danilo~P, Zhang, Liqing, Ball, Tonio,
  Schulze-Bonhage, Andreas, and Cichocki, Andrzej.
\newblock Multilinear subspace regression: An orthogonal tensor decomposition
  approach.
\newblock In \emph{NIPS}, volume 2011, pp.\  1269--1277, 2011.

\bibitem[Zhou et~al.(2013)Zhou, Li, and Zhu]{zhou2013tensor}
Zhou, Hua, Li, Lexin, and Zhu, Hongtu.
\newblock Tensor regression with applications in neuroimaging data analysis.
\newblock \emph{Journal of the American Statistical Association}, 108\penalty0
  (502):\penalty0 540--552, 2013.

\end{thebibliography}
\bibliographystyle{icml2016}
}
\end{document}